\documentclass[letterpaper, 10 pt, conference]{ieeeconf}  

\IEEEoverridecommandlockouts                              

\usepackage{amsmath,amssymb,amsfonts}
\usepackage{algorithmic}
\usepackage{graphicx}
\usepackage{textcomp}
\usepackage[sorting=none, style=ieee]{biblatex}
\usepackage{amsmath}
\usepackage{flushend}
\usepackage{booktabs}
\usepackage{makecell}
\usepackage{threeparttable}
\usepackage{multirow}
\usepackage{gensymb} 

\def\BibTeX{{\rm B\kern-.05em{\sc i\kern-.025em b}\kern-.08em
    T\kern-.1667em\lower.7ex\hbox{E}\kern-.125emX}}
\begin{document}
\title{BronchoTrack: Airway Lumen Tracking for Branch-Level Bronchoscopic Localization}
\author{Qingyao Tian, Huai Liao, Xinyan Huang, Bingyu Yang, Jinlin Wu, Jian Chen, Lujie Li, Hongbin Liu
\thanks{
Qingyao Tian, Bingyu Yang and Jian Chen are with Institute of Automation, Chinese Academy of Sciences, Beijing 100190, China, and also with the School of Artificial Intelligence, University of Chinese Academy of Sciences, Beijing 100049, China.
}
\thanks{
Huai Liao, M.D. and Xin-yan Huang, M.D. are with Department of Pulmonary and Critical Care Medicine, The First Affiliated Hospital of Sun Yat-sen University, Guangzhou, Guangdong, China.
}
\thanks{
Jinlin Wu is with Institute of Automation, Chinese Academy of Sciences, and with Centre of AI and Robotics, Hong Kong Institute of Science\& Innovation, Chinese Academy of Sciences.
}
\thanks{
Lujie Li, M.D. is with Department of Radiology, The First Affiliated Hospital, Sun Yat-sen University, Guangzhou, Guangdong, China
}
\thanks{
Corresponding author: Hongbin Liu is with Institute of Automation, Chinese Academy of Sciences, and with Centre of AI and Robotics, Hong Kong Institute of Science\& Innovation, Chinese Academy of Sciences. He is also affiliated with the School of Engineering and Imaging Sciences, King’s College London, UK. (e-mail: liuhongbin@ia.ac.cn).
}}

\maketitle

\pubid{\begin{minipage}{\textwidth}\ \\[30pt] \centering
	This work has been submitted to the IEEE for possible publication. Copyright may be transferred without notice, after which this version may no longer be accessible.
\end{minipage}}

\begin{abstract}
Localizing the bronchoscope in real time is essential for ensuring intervention quality. However, most existing methods struggle to balance between speed and generalization. To address these challenges, we present BronchoTrack, an innovative real-time framework for accurate branch-level localization, encompassing lumen detection, tracking, and airway association. To achieve real-time performance, we employ benchmark light weight detector for efficient lumen detection. We firstly introduce multi-object tracking to bronchoscopic localization, mitigating temporal confusion in lumen identification caused by rapid bronchoscope movement and complex airway structures. To ensure generalization across patient cases, we propose a training-free detection-airway association method based on a semantic airway graph that encodes the hierarchy of bronchial tree structures. Experiments on nine patient datasets demonstrate BronchoTrack's localization accuracy of 85.64\%, while accessing up to the 4th generation of airways. Furthermore, we tested BronchoTrack in an in-vivo animal study using a porcine model, where it localized the bronchoscope into the 8th generation airway successfully. Experimental evaluation underscores BronchoTrack’s real-time performance in both satisfying accuracy and generalization, demonstrating its potential for clinical applications.
\end{abstract}


\section{Introduction}

Lung cancer leads to the highest number of global cancer-related deaths \cite{siegel2023cancer}, often diagnosed at an advanced stage with a low survival rate \cite{thandra2021epidemiology}. In contrast, early diagnose significantly improves patient outcomes \cite{thandra2021epidemiology}. Bronchoscopy serves as the golden standard for accurately inspecting both central and distal airway lesions \cite{andolfi2016role}. During interventions, a flexible bronchoscope equipped with a distal camera is used to navigate to target nodules identified in pre-operative CT scans. However, due to the limited field of view, precise localization within the lung requires extensive clinical experience. 

Multiple sensing modalities, such as electromagnetic navigation \cite{sadjadi2015simultaneous}, 3-D shape sensing \cite{shi2016shape} and endoscope vision can assist bronchoscopic localization. Visually navigated bronchoscopy (VNB) requires no additional equipment, therefore is preferred as it offers cost-effectiveness and ease of setup. However, due to the inherent instability and dynamics of bronchoscope movement, as well as the complex anatomical tubular structures, current VNB methods face limitations in meeting application demands.

The primary challenge lies in striking a balance between computational speed and creating a streamlined pipeline to generalize across a diverse spectrum of patients. Current VNB methods fall into two main categories: retrieval-based \cite{zhao2019generative,sganga2019autonomous} and registration-based localization \cite{mori2002tracking,deguchi2009selective,shen2019context,banach2021visually}. Retrieval-based methods require individualized patient training. Registration-based approaches depend on iterative optimization, leading to diminished computational speed. In contrast to existing VNB methods, surgeons rely on recognizing and tracking airway lumens and bifurcations to monitor bronchoscopic motion and determine bronchoscope location within the corresponding airway branches. This observation indicates potential for improving VNB by replicating the human approach.

Motivated by above observations, this paper proposes a novel localization pipeline called BronchoTrack, that tackles the challenge of balancing speed and generalization. Inspired by the surgeon's approach to bronchoscope localization, BronchoTrack achieves localization through a combination of lumen detection and tracking, coupled with the mapping of each detected lumen to the patient's unique airway anatomy. 

To achieve real-time speed, BronchoTrack employs an efficient lightweight detector for lumen detection. BronchoTrack pioneers the integration of multi-object tracking (MOT) into bronchoscopic localization, alleviating temporal confusion in lumen identification caused by rapid bronchoscope movement and complex airway structures.

To ensure patient generalization, BronchoTrack integrates a training-free detection-airway association method using a semantic airway graph encoding bronchial tree hierarchy. This approach allows us to map observations to the airway anatomy in a single attempt, ensuring both accuracy and computational speed across diverse patients’ domain.

Additionally, BronchoTrack incorporates a closed-loop search function inspired by SLAM to enhance tracking stability. Leveraging image features encompassing a broader global context, the proposed adapted loop closure module helps to overcome the vulnerability of lumen tracking to challenges such as scene deformations, occlusions, and blurriness. Experiments on patient data and real-time animal studies confirm BronchoTrack’s potential to assist surgeons and enable autonomous driving bronchoscopy.

In summary, our contributions are as follows:

\begin{itemize}
    \item We propose a VNB pipeline for real-time bronchoscope localization without the need for patient-specific retraining, which includes an efficient lumen detector, a multi-lumen tracker, and a detection-airway association module.
    \item We pioneer the integration of multi-object tracking techniques into bronchoscope navigation to prevent misidentification of detected lumens.
    \item We employ patient-specific airway models generated from pre-operative CT scans to construct a semantic airway graph for assigning branchial labels of detected lumens. The semantic airway graph conveys the domain knowledge among patients. 
    \item Our approach is validated through in-vivo animal studies with a porcine model and extensive offline experiments on patient data, demonstrating its practical potential.
\end{itemize}

\begin{figure*}[!t]
\centerline{\includegraphics[width=\textwidth]{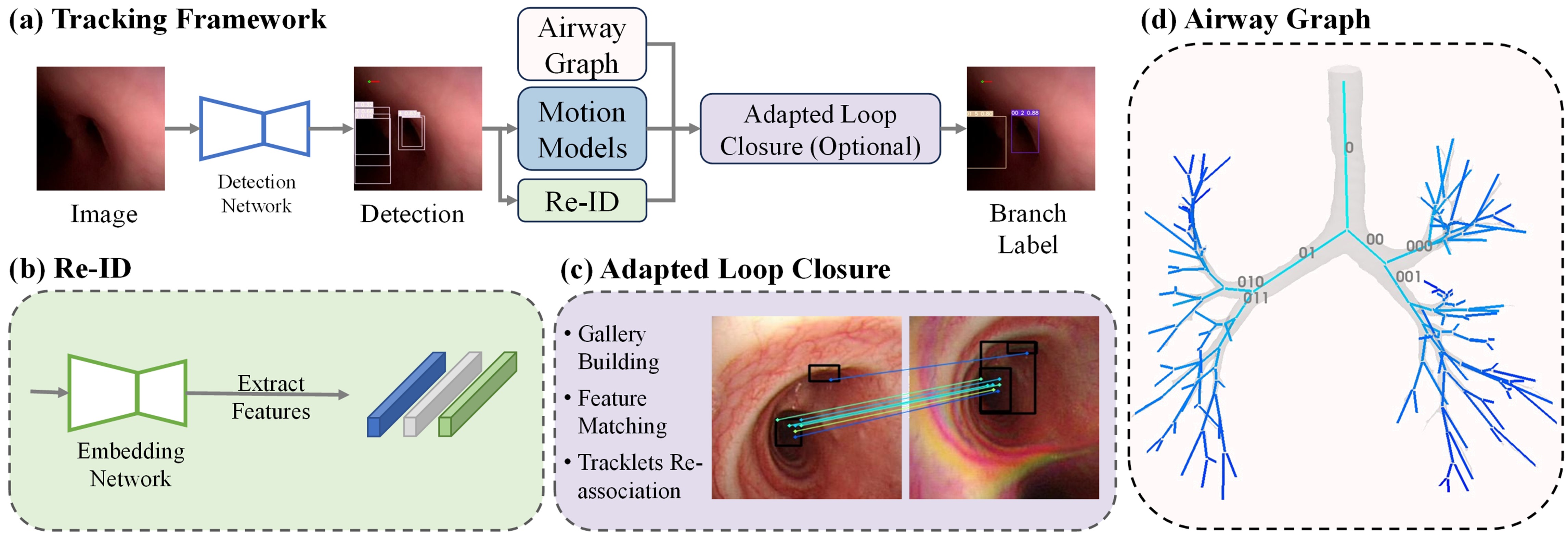}}
\caption{(a) Diagram of BronchoTrack. (b)(c)(d) are the modules of BronchoTrack. Based on motion models of tracked lumens, BronchoTrack predicts bounding box position. Combining with deep feature descriptor extracted by Re-ID branch, BronchoTrack matches new detections with previous tracklets to propagate airway labels. After building an airway subgraph with incomplete airway labels from tracklets, the labels are refined based on contextual information and anatomical constraints from pre-operative airway graph. Finally, the refined tracklets are used to generate the coarse bronchoscope localization precise to branch anatomical level. With BronchoTrack-LC extension, we search closed loops after initial tracklets association to detect and recover from tracking failure.}
\label{fig1}
\end{figure*}

\section{Related Work}

Early VNB studies focus on matching between the intensity of virtual and real bronchoscopic frames. Those methods either suffer from a slow update rate (1-2 Hz) due to constant rendering of virtual view during iteration \cite{mori2002tracking,deguchi2009selective} or require high rendering quality of the virtual images to have similar texture and illumination with real bronchoscopy frames \cite{merritt2013interactive,luo2014discriminative}, restricting their practical application.

Recently, some groups \cite{karaoglu2021adversarial,shen2019context,banach2021visually,mathew2020augmenting} have employed deep learning for bronchoscopic frames depth estimation and locate the bronchoscope by registrating estimated depth to pre-operative airway model. However, their registration efficiency raises concerns for real-time applications. Other studies \cite{ozyoruk2021endoslam,sganga2019offsetnet} investigate deep visual odometry techniques for predicting the camera motion between frames, but they encounter challenges like scale ambiguity or poor generalization. Zhao et al. \cite{zhao2019generative} attempt to address this by using auxiliary learning to train a global pose regression network \cite{valada2018deep}. Nonetheless, global pose learning is essentially image retrieval and struggles to generalize beyond its training data \cite{sattler2019understanding}. Feature based visual SLAM methods have been adopted \cite{wang2019visual,wang2020visual}, but they may face tracking failure due to the lack of features on the textureless lumen surface. Attempt has been made to navigate bronchoscope by bifurcation detection \cite{shen2017branch}, but computational speed remains a concern for practical application. 

Closely related to our work, \cite{sganga2019autonomous} proposes AirwayNet and BifurcationNet to localize the bronchoscope by estimating visible airways with their relative poses. However, AirwayNet requires individualized training for each patient, and successful navigation is limited by detection accuracy, which critically depends on comprehensive data collection. BifurcationNet does not require specific airway for training, but relies on assistance of other modalities and performs worse than AirwayNet. Those methods either demand retraining for anatomical labeling, or enroll manually designed prior assumptions to associate detections between frames and with pre-operative CT. 

In order to cope with shortcomings of coordinate level  bronchoscope localization methods, several studies take a different approach to bronchoscopic navigation, focusing on branch-level estimation through lumen detection \cite{wang2021depth,esteban2016stable,sanchez2017towards}. Branch-level localization provides sufficient guidance for surgeons and has the potential to be complementary to coordinate level estimation by providing registration search space, improving its robustness and inference speed. However, existing studies usually detect lumens by traditional methods, which heavily depend on assumptions about the airway geometry and may not operate in real-time. Moreover, lumen tracking is implemented in a naïve approach solely based on lumen position, which is unreliable due to bronchoscope dynamic. 

Our work is inspired by prior studies in branch level localization through lumen detection. However, we make significant advancements by introducing the novel framework, BronchoTrack. To the best of our knowledge, BronchoTrack is the first branch-level localization method to achieve real-time performance consistently across datasets with high accuracy. Additionally, it is the first VNB method to be validated in real-time during a robot-assisted animal experiment.

\section{Method}
In this section, we introduce a lumen tracking pipeline termed as BronchoTrack, aiming at bronchoscope localization and navigation. The diagram of our algorithm is presented in Fig. \ref{fig1}. BronchoTrack is a three-branch framework encompassing lumen detection, tracking, and bronchoscope localization. The tracking module comprises motion models and appearance models for associating detections with tracklets. Localization is achieved through an airway association module, matching activated tracklets with pre-operative CT scan. In this context, a tracklet $T$, which is a sequence of detections of an object in consecutive frames, can be represented as a tuple: 

\begin{equation}\label{eq:example}
    T=\left\{I D,\left\{ind _{start}, ind_{end}\right\},\left\{b_i, b_{i+1}, \ldots, b_j\right\}\right\},
\end{equation}

\noindent which includes the unique identifier for the tracklet $ID$, start frame index of the tracklet $ind_{start}$, end frame index of the tracklet  $ind_{end}$, and the sequence of bounding boxes $\left\{b_i, b_{i+1}, \ldots, b_j\right\}$ representing the object's trajectory.

BronchoTrack-LC is a BronchoTrack extension including an adapted loop closure module for tracking recovery, to deal with rapid bronchoscope movement and occlusion. 

Before intervention, airway segmentation \cite{zheng2021alleviating} and skeletonization \cite{lee1994building} are performed with patient specific CT scan. The central axis of the bronchus is further divided by their branching, and then start and end points of each segment are preserved to form a topological tree as the airway graph. We transform the airway graph into a standard coordinate where the y-axis aligns with the direction of trachea, the x-axis lies in the plane formed by the origin and the end of the left and right main bronchus, and z-axis is orthogonal to both x-axis and y-axis. Denote the standard airway graph as $M$. Each branch of $M$ is uniquely labeled and its parent and child branches are recorded (Fig. \ref{fig1} (d)). The goal of BronchoTrack is matching tracklets with new detections and associating tracklets with corresponding branch labels in airway graph. Note that due to possible missing association, the relationship between branchial labels and tracklets is a one-to-many mapping.

\subsection{Lumen Detection}
Real-time lumen detection is crucial for our tracking and localization pipeline. We have integrated the high-performance YOLOv7 detector \cite{wang2023yolov7} for computational speed. This includes the use of Extended Efficient Layer Aggregation (E-ELAN) for faster convolutional layers, trainable ``bag-of-freebies" optimization modules, such as RepConv without identity connection, and coarse-to-fine lead guided label assignment for improved training \cite{lee2015deeply}. YOLOv7's parameter and computation reductions of 40\% and 50\%, respectively, enhance BronchoTrack's inference speed and detection accuracy.

\subsection{Multi-lumen Tracking}

\textbf{Motion Models.} The accuracy of MOT is highly dependent on the overlap between the bounding boxes of predicted tracklets and the detected objects, because assumption is made that detected objects with larger IoU with the predicted bounding boxes of previous tracks should be associated as the same tracklets. However, with the rapid movement of endoscope, the bounding box location could shift dramatically in the scene, leading to possible violation of this assumption, which would result in false matching of detections.

BronchoTrack utilizes motion pattern to predict temporal-spatial changes of lumen’s trajectories in the image plane. In addition, it allows for tracking extrapolations in case of detector failures. We adopt the Kalman filter \cite{kalman1960new} with a constant-velocity model to predict the tracklets’ motion in the image plane, as in \cite{du2023strongsort,wojke2017simple,han2022mat,zhang2022bytetrack}. At each time step, the current state vector of tracklets, $\boldsymbol{x}_k=$ $\left[x_c, y_c, h, a, \dot{x}_c, \dot{y}_c, \dot{h}_c\right]$, is predicted before associating new detections with tracklets, where $\left(x_c, y_c\right)$ represent the 2D object center coordinates in the image plane. $h$ and $a$ are the bounding box height and aspect ratio. $\left(\dot{x}_c, \dot{y}_c\right)$ and $\dot{h}_c$ are the velocities of object center and bounding box height, respectively. 

By predicting tracklets’ position in the current frame, we can reasonably assume that larger overlapping between the predicted bounding boxes of tracklets and the detected objects suggests greater probability of matching. Therefore, we calculate the motion matching cost between the $i$-th tracklet and the $j$-th detection as follows:

\begin{equation}\label{eq:example}
    C_m(i, j)=1-\text{IoU}\left(\boldsymbol{x}_i, \boldsymbol{z}_j\right),
\end{equation}

\noindent where $\text{IoU}(\cdot)$ stands for intersection over union between two bounding boxes. $\boldsymbol{x}_i$ is the predicted state of the $i$-th tracklet and $\boldsymbol{z}_j$ represents bounding box of the $j$-th detection.

\vspace{0.3cm}
\textbf{Re-ID.} Due to the dynamic of the bronchoscope and potential occlusions in the field of view, relying solely on the Kalman filter for tracking becomes challenging over extended durations. To address this issue, re-identifying objects using deep appearance cues \cite{wojke2017simple,du2023strongsort,zhang2022bytetrack,aharon2022bot} has been proposed to mitigate these challenges. 

To preserve long-term feature, we apply an exponential moving average update method to the appearance embedding $e_i^t$ of the $i$-th tracklet at time step $t$, following the approach introduced in \cite{wang2020towards}:

\begin{equation}\label{eq:example}
    e_i^t=\alpha e_i^{t-1}+(1-\alpha) f_i^t,
\end{equation}

In this context, $f_i^t$ represents the current appearance embedding for the matched detection, with a momentum of $\alpha=0.9$ set for exponential moving average. To associate tracklets with current detections, the Re-ID network calculates feature embeddings for each detection box frame crop and assesses appearance matching cost between the $i$-th tracklet and the $j$-th detection by the following approach:

\begin{equation}\label{eq:example}
    C_a(i, j)=1-\left(f_j^t\right)^T e_i^{t-1}.
\end{equation}

We modify matching strategy from \cite{zhang2022bytetrack}. Initially, new detections are categorized into high and low confidence candidates. Candidate tracklets are filtered if their airway labels are distant (more than three generations away) from estimated bronchoscope location of the previous frame. For high confidence detections, we use both appearance and motion distance criteria for matching, represented by:

\begin{equation}\label{eq:example}
    C(i, j)=\lambda C_a(i, j)+(1-\lambda) C_m(i, j),
\end{equation}

In our experiment, we set the weight factor $\lambda$ to 0.5.

Forming the distance matrix between each tracklet and high confidence detection, the high-cost candidates are gated, and the remaining are matched by Hungarian Algorithm \cite{kuhn1955hungarian}. Low confidence detections together with unmatched high confidence detections are matched with unmatched tracklets again using only motion distance. Finally, unmatched high confidence detections are added as new tracklets, while low confidence detections are treated as background and removed.

\begin{figure*}[!t]
\centerline{\includegraphics[width=\textwidth]{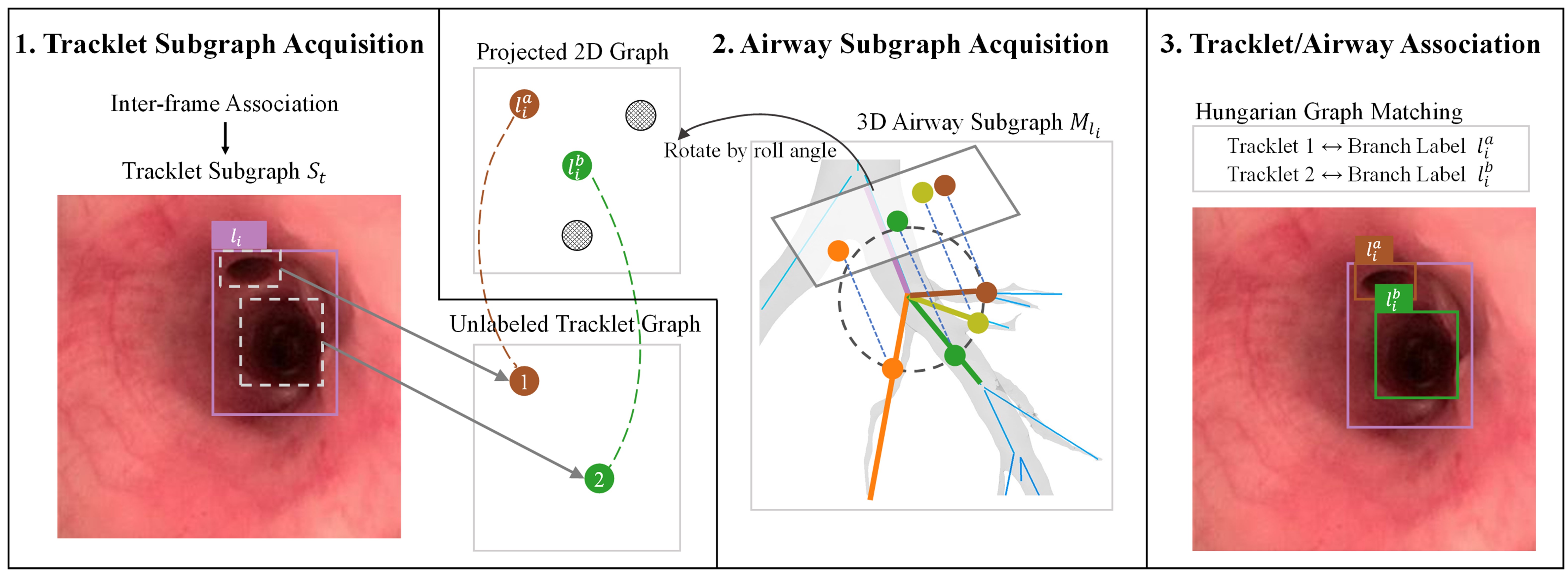}}
\caption{Illustration of BronchoTrack's airway association. Here, we exemplify the labeling of observed child branches under $l_i$. 1. Leveraging inter-frame association, the label $l_i$ is initially assigned. 2. We determine observed branches by considering the likelihood of observing a child branch of $l_i$ based on a negative correlation between its angle against $l_i$. Subsequently, we generate a 2D graph of  $l_i$'s child branches projected from the 3D airway graph. The 2D graph is further rotated based on the estimated bronchoscope roll angle. 3. Ultimately, the Hungarian Algorithm is employed to match the two graphs to associate tracklets with airway branches.}
\label{fig2}
\end{figure*}

\subsection{Airway Graph Association}
After associating current frame detections with tracklets, the next step involves determining anatomical labels for observed tracklets, predicting the bronchoscope's branch-level position. To achieve this, we introduce the airway association module, which uses the semantic airway graph to incorporate domain knowledge gathered from multiple patients.

At time step $t$, we create a subgraph $S^t$ of observed lumens representing their hierarchy based on detection bounding box intersections and inclusions. This subgraph, in combination with the airway graph $M$, is used to infer airway branch labels for tracklets.

\vspace{0.3cm}
\textbf{Initialization.} BronchoTrack initializes before carina, where the two detected lumens of primary level are identified as left and right main bronchus respectively. The bronchoscope roll angle is initiated according to the position of primary level bounding box centers by:

\begin{equation}\label{eq:example}
    \mathrm{roll}_0=\arccos \left(\frac{[1,0] \cdot\left(c_r-c_l\right)}{\left\|\left(c_r-c_l\right)\right\|}\right),
\end{equation}

\noindent where $c_l$ and $c_r$ are the position of bounding boxes of the left and right main bronchus in the image plane.

This approximate roll angle is updated in every new frame and is used to decide the mapping from sibling bounding boxes to airway branches.

\vspace{0.3cm}
\textbf{Gallery Building.} We build a gallery of visited bronchi, denoted as $G = \left\{ {G_{l_{0}}:\left( {T_{0},\mathrm{roll}_{0}} \right),\ldots,{G}_{l_{i}}:\left( {T_{i},\mathrm{roll}_{i}} \right),\ldots,}\right.$ $\left.{G_{l_{n}}:\left( {T_{n},\mathrm{roll}_{n}} \right)} \right\}$, which contains the record $G_{l_i}$ for each traversed airway branch $l_i$. Each individual record, such as $G_{l_i}$, documents all the identified tracklets $T_i=\left\{T_{i 1}, T_{i 2}, \ldots\right\}$ and the roll angle $\mathrm{roll}_i$ while inspecting branch $l_i$.

During the intervention, a new record is added to the gallery $G$ when the bronchoscope accesses an unexplored branch. $G_{l_i}$ is updated by a new set of tracklets and bronchoscope roll whenever the number of observed lumens at the present surpasses the existing gallery records for the corresponding branch $l_i$. This count of detected lumens is used as a criterion to determine whether the current frame provides a more comprehensive hierarchy of the branch $l_i$.

\vspace{0.3cm}
\textbf{Bronchoscope Roll Estimation.} At every timestep, we sort the two oldest tracklets and retrieve their previous observations from the gallery, specifically denoting the corresponding record as $G_{l_m}=\left(T_m, \mathrm{roll}_m\right)$. The current bronchoscope roll can be estimated in a preliminary manner by:

\begin{equation}\label{eq:example}
  \mathrm{roll}_t=\mathrm{roll}_m+\arccos \left(\frac{\left(c_1^m-c_2^m\right) \cdot\left(c_1^t-c_2^t\right)}{\left\|\left(c_1^m-c_2^m\right)\right\|\left\|\left(c_1^t-c_2^t\right)\right\|}\right),
\end{equation}

\noindent where $c_1^t$ and $c_2^t$ denote the bounding box center positions of the two oldest tracklets at current frame, while $c_1^m$ and $c_2^m$ are their previous center positions recorded at $T_m$.

\vspace{0.3cm}
\textbf{Intra-frame Association.} Inter-frame association of tracklets extends branch labels into the current frame. New branches or unassociated detections remain unlabeled during this stage and are assigned labels during intra-frame association.

Starting with the tracklet subgraph $S^t$ containing incomplete branchial labels, we iterate over labeled detections by track age to propagate labels to unlabeled sections of $S^t$. In each iteration, the process begins with the reference detection $b_i^t$, recognized as branch $l_i$. If $b_i^t$'s parent lumen lacks a label, it is labeled as the parent branch of $l_i$, as indicated in airway graph $M$. Additionally, if any child lumens linked to $b_i^t$ are unlabeled, all child lumens are labeled using subgraph $M_{l_i}$ originating from branch $l_i$. Similarly, if any sibling branches of $b_i^t$ are unlabeled, all sibling branches are labeled using subgraph $M_{l_i^p}$ originating from $l_i$'s parent branch $l_i^p$. This process ensures consistent labeling of the entire subgraph.

Fig. \ref{fig2} provides the outline of mapping from child lumens of $b_i^t$ to branches in $M_{l_i}$. Since not all child branches of $l_i$ are observed at time $t$, the association between tracklets and labels commences by identifying the observed branches. We assess the likelihood of observing a child branch $l_i^j$ (where $j = 1, \ldots, m$, and $m$ represents the number of $l_i^j$'s child branches) of branch $l_i$ based on a negative correlation between the intersection angle of $l_i$ and its child branches.

To bridge the gap between child branches in the 3D airway graph and their 2D image plane detections, we create a 2D graph from the 3D airway subgraph $M_{l_i}$. This is achieved by truncating $l_i$'s child branches beyond a certain distance and projecting them onto $l_i$'s tangent plane. We then rotate the 2D graph based on the estimated bronchoscope roll angle. However, due to an unknown transformation between image plane and 2D airway graph coordinates, the correspondence between detections and branches cannot be established directly. To address this, we treat detected lumens and candidate branches as separate graphs, turning the association problem into a general graph matching. In this context, we use the Hungarian Algorithm \cite{kuhn1955hungarian} to associate tracklets with branches in the airway graphs. Similar methods are applied for the association between sibling lumens of $b_i^t$ and branches in $M_{l_i^p}$.

\vspace{0.3cm}
\textbf{Localization.} To infer bronchoscope position from the labeled subgraph $S^t$ at time step $t$, we adopt a voting strategy to deal with false annotating. Denote the number of detected lumens of primary level with $n$, for each branch with airway label $l_i$ at level $k$, it votes for location of:

\begin{equation}\label{eq:example}
  loc=\left\{\begin{array}{c}
g^{k-1}\left(l_i\right), \quad \text { if } n=1 \\
g^k\left(l_i\right), \quad \text { if } n>1
\end{array}\right.,
\end{equation}

\noindent where $g^k(\cdot)$ denotes the $k$-th level above branch. The most voted branch is considered as the bronchoscope location.

\begin{figure*}[!t]
\centerline{\includegraphics[width=\textwidth]{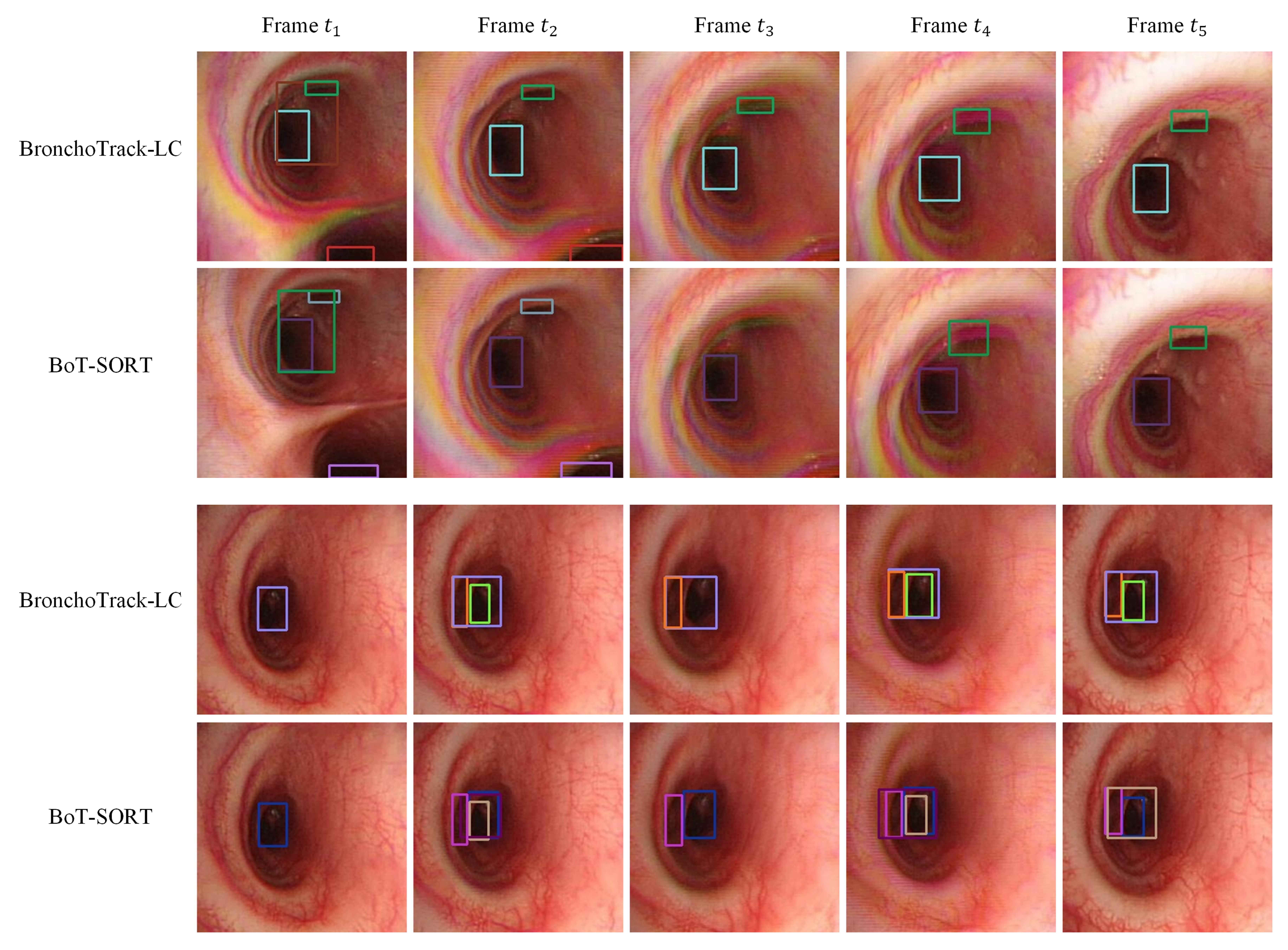}}
\caption{Visualization results of BronchoTrack-LC compared with BoT-SORT. We demonstrate BronchoTrack-LC's ability to handle challenging cases, such as motion blur and occluded lumens, by selecting sequences from the detector's test set. The identical box color signifies the same identity.}
\label{fig3}
\end{figure*}

\subsection{Adapted Loop Closure for Tracking Recovery}
Existing MOT studies predominantly emphasize applications involving high frame rates and cameras with minimal movement. However, due to the irregular movements of bronchoscope, motion prediction becomes more challenging with regard to bronchoscopic videos. Furthermore, the reliability of appearance features can be compromised due to issues such as motion blurs and artifacts.

We introduce adapted loop closure techniques called BronchoTrack-LC, inspired by loop detection in SLAM systems \cite{mur2015orb,mur2017orb}. We expand the airway gallery $G$ with image frames to enhance tracking. Loop closure searches for loops when the bronchoscope enters a new branch, designating the corresponding frame as a keyframe. To address textureless airway lumens, we employ LoFTR \cite{sun2021loftr}, a dense feature matching network, for matching between keyframes in the gallery and new keyframes. A new keyframe is considered matched with gallery frames if more than $\eta$ matching keypoint pairs are detected. To maintain real-time performance of BronchoTrack-LC, the new keyframe is compared with the $\lambda$ most recently updated gallery records. We use $\eta=100$ and $\lambda=1$ in our experiment. If a loop is detected, we recompute tracklet associations between the matched keyframe and the current frame. Otherwise, we insert the new keyframe of the newly visited branch into $G$.

Notably, while Re-ID also utilizes appearance features for detection identification, it focuses on local appearances. In contrast, loop closure uses explicit feature matching, establishing pixel-wise dense matches with a larger global context of keyframes to overcome vulnerability to deformed, occluded, and blurred lumens.

\section{Experiments}
\subsection{Implementation}

The detector is trained on patient data from the First Affiliated Hospital, Sun Yat-sen University, approved by the IEC for Clinical Research. The training dataset consists of 7644 manually labeled bronchoscopic video frames from ten patients, all resized to 256×256 for detector network training and testing. The same detector weight is used for both patient data and the animal experiment.

The Re-ID module employs a ResNet50 architecture for image classification/retrieval training. For data preparation, cropped detection boxes from the same lumen are treated as one category, containing 3630 cropped images of lumen bounding boxes resized to 128×128. The network is trained using softmax classification loss. During inference, the final fully connected layers for predicting lumen IDs are omitted, and the CNN computes the appearance embedding of input frame crops.

In both patient and in-vivo data, we use a detection threshold of 0.1 and IoU thresholds of 0.6 and 0.7. A relaxed IoU threshold accounts for overlapping lumen hierarchies. We construct a local airway subgraph based on the overlap of detection boxes, pruning redundant parent branches and isolated child branches with large IoU with their parents. For tracklet association, we use a matching threshold of 0.4 for high-score detections. If there are matched pairs after the initial association, we use a low score matching threshold of 0.7; otherwise, the threshold is set to 0.9.

\begin{table}[t]
  \centering
  \footnotesize
  \caption{Detection metrics.}
    \begin{tabular}{ccccc}
    \toprule
          & Precision & Recall & AP@0.5 & mAP@.5:.95 \\
    \midrule
    Porcine   & 0.788 & 0.785 & 0.781 & 0.399 \\
    Human\_train & 0.901 & 0.911 & 0.932 & 0.779 \\
    Human\_test & 0.878 & 0.896 & 0.902 & 0.725 \\
    \bottomrule
    \end{tabular}%
  \label{table1}%
\end{table}%

\begin{figure*}[!t]
\centerline{\includegraphics[width=\textwidth]{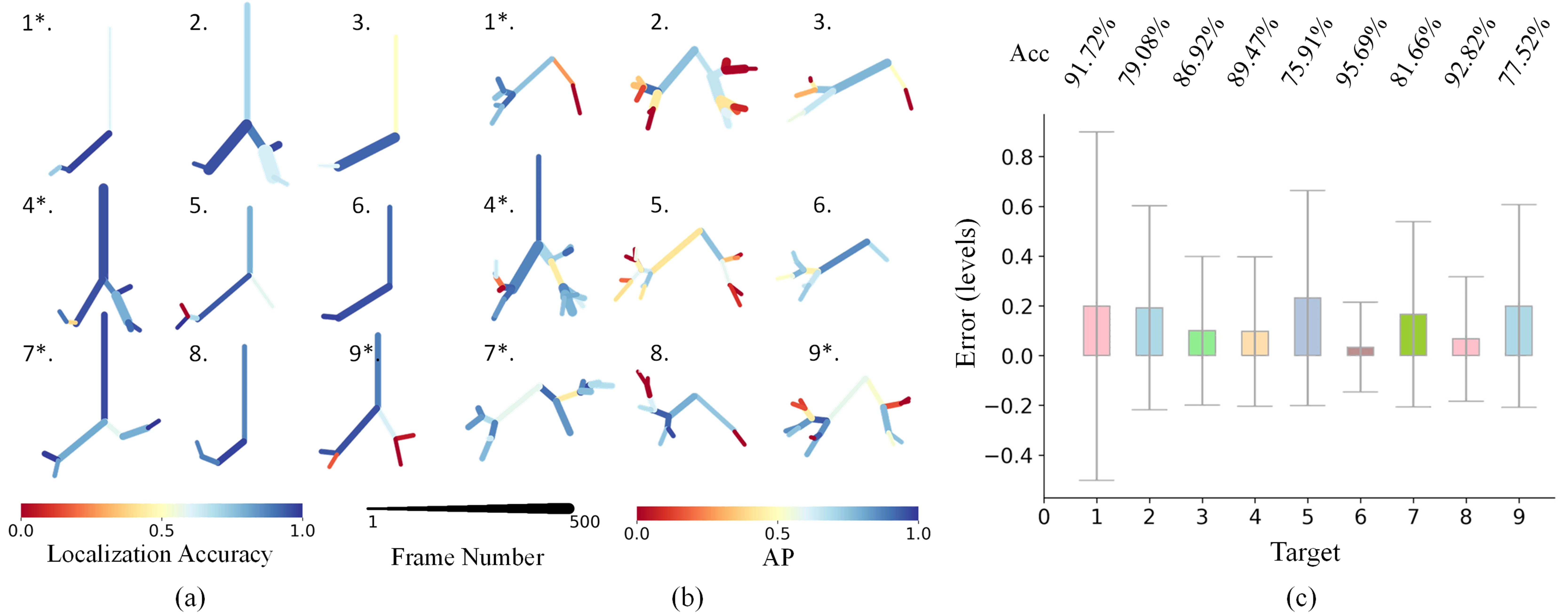}}
\caption{BronchoTrack-LC testing results in human data. (a) Localization accuracy at each visited branch. (b) Average precision of identifying each visible branch as its corresponding branch label. (c) Average localization error among each trajectory. Localization accuracy of each case is listed above the graph. * denotes training cases for detector.}
\label{fig4}
\end{figure*}

\begin{table*}[t]
  \centering
  \footnotesize
  \caption{Performance of lumen tracking on patient and porcine data.}
  \resizebox{\textwidth}{!}{
    \begin{tabular}{c|cccccc|cccccc|c}
    \hline
    \multirow{2}{*}{\textbf{Trackers}} & \multicolumn{6}{c|}{\textbf{Patient}}         & \multicolumn{6}{c|}{\textbf{Porcine}}         & \multirow{2}{*}{\textbf{FPS↑}} \\
    \cline{2-13}
    & \textbf{MOTA↑} & \textbf{IDF1↑} & \textbf{HOTA↑} & \textbf{FP↓} & \textbf{FN↓} & \textbf{IDs/GT\_IDs} & \textbf{MOTA↑} & \textbf{IDF1↑} & \textbf{HOTA↑} & \textbf{FP↓} & \textbf{FN↓} & \textbf{IDs / GT\_IDs} &  \\
    \hline
    \textbf{Sort} & 29.893 & 22.311 & 23.502 & 7033 & 8281 & 2125/146 & 57.115 & 44.071 & 41.792 & 9789 & 8384 & 925/81 & \textbf{191.5} \\
    \textbf{DeepSort} & 18.399 & 32.003 & 28.159 & 8712 & 6824 & 679/146 & 46.346 & 49.732 & 41.124 & 9870 & 7133 & 271/81 & 77.8 \\
    \textbf{ByteTrack} & 30.57 & 36.407 & 30.838 & 8842 & 6180 & 78/146 & 57.476 & 55.602 & 45.116 & 9052 & 6073 & 63/81 & 80.4 \\
    \textbf{StrongSort} & 28.138 & 39.893 & 33.342 & 8403 & 5775 & 91/146 & 50.521 & 52.731 & 43.415 & 9533 & 6565 & 67/81 & 68.7 \\
    \textbf{BotSort} & 39.613 & 40.111 & 36.614 & 7625 & 5938 & 601/146 & 60.145 & 56.727 & 49.088 & 8566 & 5998 & 258/81 & 69.8 \\
    \textbf{BronchoTrack w/o graph} & 51.536 & 55.414 & 45.497 & 4215 & 4848 & 515/146 & 62.313 & 61.468 & 50.179 & 4958 & 6447 & 1146/81 & 67.1 \\
    \textbf{BronchoTrack(ours)} & 51.583 & 56.228 & 45.947 & 4133 & 4765 & 502/146 & \textbf{66.173} & \textbf{75.839} & \textbf{56.677} & \textbf{2260} & \textbf{4669} & \textbf{89/81} & 51.8 \\
    \textbf{BronchoTrack-LC(ours)} & \textbf{59.061} & \textbf{74.246} & \textbf{55.632} & \textbf{2284} & \textbf{2944} & \textbf{153/146} & 64.777 & 75.712 & 56.553 & 2942 & 4283 & 98/81 & 32.8 \\
    \hline
    \end{tabular}}
  \label{table2}
\end{table*}

\subsection{Setup}
We validate our method through both offline experiments on patient data and an in-vivo experiment using a porcine model to simulate clinical scenarios and assess performance in deeper airway generations.

For the offline experiments, we evaluate BronchoTrack using nine human cases. Four of these cases are included in the training dataset for detector and Re-ID models, demonstrating performance on unseen data. The patient data are from regular bronchoscopy inspections with Olympus bronchoscopes, recording at approximately 15fps. Fragments with poor camera visibility due to fluid occlusion are manually removed, preserving long trajectories starting before the carinas for testing. The remaining fragments involve inspection into the 3-4 airway generations.

The animal experiment takes place at the Animal Experiment Center of the First Affiliated Hospital of Sun Yat-sen University under IRB approval. A bronchoscope mounted on a bronchoscopic robot captures video at 30fps. BronchoTrack's tracking and localization results are recorded during the animal trial and later evaluated using manually labeled bronchoscopic frames as ground truth. For real-time in-vivo experiments, we use BronchoTrack due to its faster inference speed. We exclude trials with unclear or incomplete airway views, resulting in the inclusion of six insertion trajectories for evaluation.

All training and testing are conducted using the PyTorch framework on an NVIDIA RTX3090 GPU. Data annotations are validated by experienced surgeons.

\subsection{Metrics}
We evaluate BronchoTrack and BronchoTrack-LC by comparing them with baseline and state-of-the-art trackers. We use metrics such as MOTA, FP, FN, IDs, IDF, and HOTA to assess performance, with IDF being particularly crucial for accurate localization.

Additionally, we evaluate BronchoTrack's localization performance in patient data and the in-vivo experiment, reporting mean accuracy and error. We also assess localization accuracy for individual airway branches, highlighting performance across various airway generations. Furthermore, we evaluate its AP in detecting and identifying visible lumens and their corresponding branch labels.

We conduct ablation studies to analyze the impact of motion models, Re-ID, airway association, and loop closure on tracking and localization performance using both patient and porcine data.

Notably, we do not claim any algorithmic innovation in the detector. The precision, recall and mAP of the detector in patient and porcine data are reported solely for reference in the future studies (Table \ref{table1}).

\section{Results}
\subsection{Tracking Benchmark Evaluation}

We compare BronchoTrack and BronchoTrack-LC with the state-of-the-art trackers on recorded porcine data and patient data. Detectors and Re-ID models in benchmark trackers are replaced with BronchoTrack’s detector and Re-ID models. For comparison purposes, we introduce BronchoTrack without airway graph association, using the same amount of data as the benchmark methods.

\textbf{Patient Data}. BronchoTrack-LC ranks first in all metrics in patient data, improves more significantly compared to porcine data, outperforming second-performance benchmark tracker by +19.5 MOTA, +34.14 IDF and +19.01 HOTA (Table \ref{table2}). The large improvement in IDF supports more accurate branch level localization. In Fig. \ref{fig3}, we present visualization results that demonstrate BronchoTrack-LC's superior performance compared to the best performance benchmark tracker, BoT-SORT, in handling challenging tracking scenarios.

\textbf{Porcine Data}. BronchoTrack significantly surpasses the second-best benchmark tracker, achieving a substantial increase in MOTA (+4.27), IDF (+18.23), and HOTA (+7.31), all while maintaining real-time performance (Table \ref{table2}). Even without the airway association module, BronchoTrack still outperforms benchmark trackers. Note that as we rely on airway association to link detections, we adopt more stringent matching rules for tracklets, resulting in large IDs for BronchoTrack without airway association. BronchoTrack-LC closely follows BronchoTrack and achieves comparable tracking performance.

\begin{figure*}[t]
\centerline{\includegraphics[width=\textwidth]{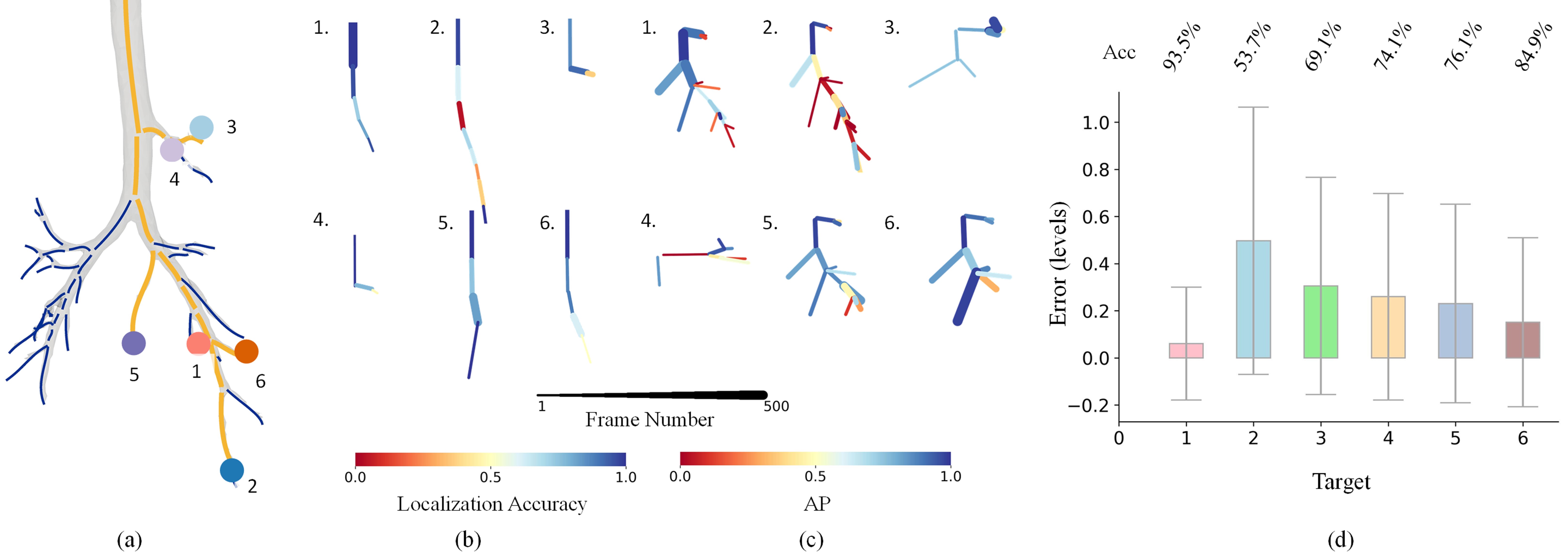}}
\caption{BronchoTrack testing results during animal experiment using porcine model. (a) the targets and driving path of each trial. (b) Localization accuracy at each visited branch. (c) Average precision of identifying each visible branch as its corresponding branch label. (d) Average localization error among each trajectory. Localization accuracy of each trajectory is listed above the graph. }
\label{fig5}
\end{figure*}

\begin{table*}[t]
  \centering
  \footnotesize 
  \caption{Results for ablation studies.}
    \begin{tabular}{@{}c|cccc|cccc|cccc@{}}
    \hline
    \multirow{2}{*}{\textbf{Methods}} & \multicolumn{4}{c|}{\textbf{Modules}} & \multicolumn{4}{c|}{\textbf{Patient}} & \multicolumn{4}{c}{\textbf{Porcine}} \\
\cline{2-13}    \multicolumn{1}{c|}{} & \textbf{KF} & \textbf{Re-ID} & \textbf{Graph} & \textbf{LC} & \textbf{MOTA↑} & \textbf{IDF1↑} & \textbf{HOTA↑} & \textbf{Loc Acc↑} & \textbf{MOTA↑} & \textbf{IDF1↑} & \textbf{HOTA↑} & \textbf{Loc Acc↑} \\
    \hline
    \textbf{BronchoTrack w/o KF} &       & \checkmark     & \checkmark     &       & 45.517 & 58.191 & 46.213 & 50.17\% & 23.694 & 41.77 & 33.543 & 47.10\% \\
    \textbf{BronchoTrack w/o Re-ID} & \checkmark &       & \checkmark     &       & 22.959 & 31.273 & 26.276 & 29.61\% & 34.174 & 40.315 & 32.314 & 27.27\% \\
    \textbf{BronchoTrack w/o Graph} & \checkmark & \checkmark     &       &       & 51.536 & 55.414 & 45.497 & -     & 62.313 & 61.468 & 50.179 & - \\
    \textbf{BronchoTrack(ours)} & \checkmark & \checkmark     & \checkmark     &       & 53.596 & 59.546 & 46.602 & 50.23\% & \textbf{66.173} & \textbf{75.839} & \textbf{56.677} & 79.11\% \\
    \textbf{BronchoTrack-LC(ours)} & \checkmark & \checkmark & \checkmark & \checkmark & \textbf{59.061} & \textbf{74.246} & \textbf{55.632} & \textbf{83.96\%} & 64.777 & 75.712 & 56.553 & \textbf{83.86\%} \\
    \hline
    \end{tabular}%
    \begin{tablenotes} 
            \item * Loc Acc denotes localization accuracy. KF represents Kalman filter. Graph stands for airway graph association module. LC represents loop closure. 
    \end{tablenotes}
  \label{table3}%
\end{table*}%

\subsection{Localization Performance}

\textbf{Patient Data.} Fig. \ref{fig4} demonstrates the localization performance of BronchoTrack-LC in patient data. BronchoTrack-LC achieved an average localization accuracy of 85.64\% across all frames in nine patient cases. Identification accuracy for higher-generation branches gradually diminishes, resulting in decreased localization performance for these branches, especially in cases not seen during training. We have observed a decline in performance for BronchoTrack when applied to patient data  (Table \ref{table3}). This can be attributed to the lower recording frame rate of patient data.

\textbf{Porcine Data.} Fig. \ref{fig5} displays the path taken to reach the targets. BronchoTrack demonstrated a commendable performance during real-time testing, with an average localization accuracy of 79.11\%. In trajectory No. 2, BronchoTrack lost track at the middle of the path, and recovered by airway association. Meanwhile, BronchoTrack-LC continues to lead in porcine data localization, obtaining average accuracy of 83.86\% (Table \ref{table3}). 

\subsection{Ablation Studies}

We evaluate different combinations of tracking and localization modules and compare their performance. As shown in Table \ref{table3}, BronchoTrack and BronchoTrack-LC outperform others in tracking metrics in porcine and patient data respectively. BronchoTrack-LC ranks first for both porcine and patient data on localization accuracy.

\section{Discussion}
Vision-based bronchoscope localization presents challenges due to the dynamic nature of bronchoscope movement and the complexities of airway structures. Our solution focuses on achieving a balance between efficiency and accuracy in real-time bronchoscope localization. We have developed a novel pipeline that includes an efficient lumen detector, a multi-lumen tracker, and a detection-airway association module. This not only enhances computational efficiency but also ensures robust performance across various patient cases.

\begin{figure}[!t]
\centerline{\includegraphics[width=\columnwidth]{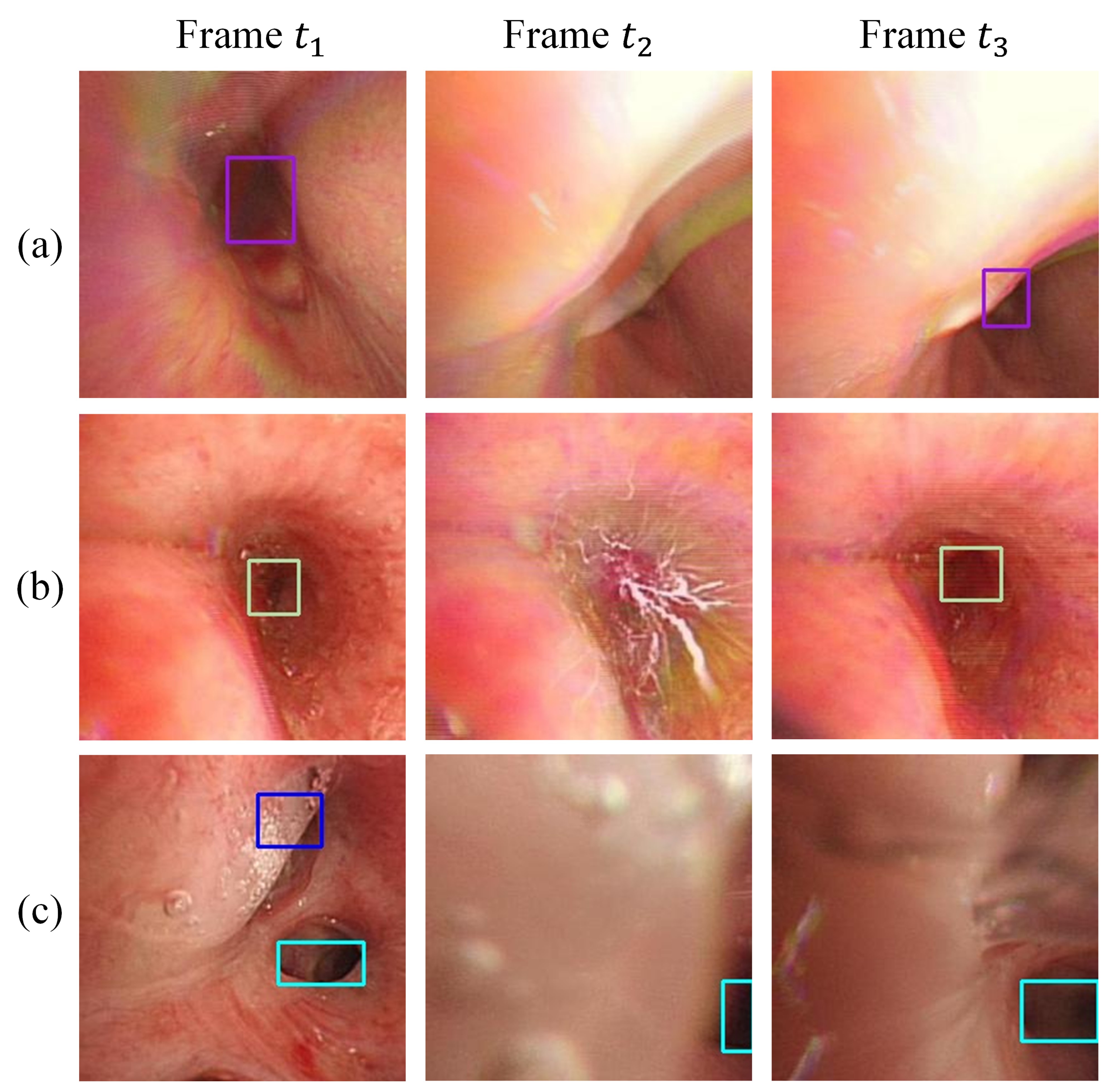}}
\caption{Examples of BronchoTrack tracking with artifacts and false lumen detections. (a) demonstrates BronchoTrack's recovery from a false negative lumen detection caused by motion blur in the video frame. (b) shows BronchoTrack recovering when lumen visibility is obstructed by liquid. (c) shows BronchoTrack's robust tracking in the presence of severe occlusions.}
\label{fig6}
\end{figure}

Lumen tracking within the airway has been explored in previous research using the Kalman filter \cite{sanchez2017towards,shen2017branch}, but relying solely on it can lead to challenges such as misidentification of detected lumens, particularly in scenarios with rapid bronchoscope movement. Our ablation study shows that retaining motion models within the tracking module results in a significant decline in tracking and localization performance. By incorporating Re-ID techniques from MOT and introducing a loop closure module, we effectively address these challenges, resulting in a more robust and dependable bronchoscope localization system, particularly in challenging scenarios like low frame rate videos (see Fig. \ref{fig6}).

Generalizing bronchoscopic navigation algorithms across different patients has been challenging due to domain gaps between cases. Our approach overcomes this challenge by embedding airway hierarchy into a semantic graph, effectively transferring domain knowledge and ensuring robustness and efficiency in a training-free manner. The semantic graph association not only enhances localization but also significantly improves tracking performance.

Our work is aligned with recent research on image-based branch-level tracking for bronchoscopy. It distinguishes itself by accommodating rotations of the bronchoscope, improving clinical alignment, and maintaining real-time performance. Compared to existing studies, we have conducted comprehensive evaluations through both offline experiments using recorded patient sequences and online assessments in an animal experiment. It represents a pioneering effort as the first visually guided bronchoscope framework with rigorous quantitative evaluations in an online setting. 

Despite its real-time performance and accuracy, BronchoTrack still has several limitations. First, tracking may fail when the bronchoscope approaches the lumen wall and enters a different branch. Notably, such situations are infrequent in human airway inspection procedures because the bronchoscope typically maneuvers along the airway's centerline to ensure an adequate field of view for surgeons. Secondly, closed-loop searching can be time-consuming. Implementing a more efficient feature matching scheme has the potential to further enhance processing speed. Thirdly, future research must involve training and testing with an even larger dataset to further validate BronchoTrack's performance.

Although BronchoTrack only provides branch-level localization, we do expect it to be complementary to existing coordinate-level localization methods. One possible strategy to incorporate branch-level estimation into coordinate-level localization includes using the tracked branch level as the initialization and boundary of registration between bronchoscopic frames and the pre-operative airway model.

\section{Conclusion}
This study introduces BronchoTrack, a branch-level bronchoscopic localization framework addressing the trade-off between speed and generalization. BronchoTrack comprises lumen detection, tracking, and airway association. For real-time performance, we employ a lightweight detector for efficient lumen detection. We introduce multi-object tracking to enhance lumen identification. To promote generalization across patient cases, we propose a training-free detection-airway association method based on a semantic airway graph encoding bronchial tree hierarchy. BronchoTrack is validated in nine patient cases, and comparative analysis with existing benchmarks is conducted. Notably, BronchoTrack is the first image-based bronchoscopic localization method quantitatively evaluated in real-time in-vivo experiments using a porcine model. Our experiments demonstrate BronchoTrack's capability for real-time bronchoscope localization across diverse cases without requiring patient-specific retraining.

\AtNextBibliography{\small}
\printbibliography
\end{document}